\documentclass{llncs}
\usepackage{epsfig}
\usepackage{fancyhdr}
\begin{document}
\title{Shortest Paths in HSI Space for Color Texture Classification}
\titlerunning {Color Texture Classification in HSI Color Space}
\toctitle {Your changed title for the table of contents}
\author{ Mingxin Jin\inst{1}, Yongsheng Dong \inst{1}$^{*}$, Lintao Zheng\inst{1}, Lingfei Liang\inst{1},\\
 Tianyu Wang\inst{1}, Hongyan zhang\inst{1}}
\institute{School of Information Engineering, Henan University of Science and Technology, Luoyang, China\\ \email{jinmingxin0501@163.com} \\\email{dongyongsheng98@163.com}}
\maketitle

\pagestyle{fancy}
\fancyhead[RO]{Color Texture Classification in HSI Color Space\quad \thepage }
\fancyhead[LE]{\thepage\quad ICIS2018 }
%\fancyhead[LE,RO]{\thepage~}
\begin{abstract}
Color texture representation is an important step in the task of texture classification. Shortest paths was used to extract color texture features from RGB and HSV color spaces. In this paper, we propose to use shortest paths in the HSI space to build a texture representation for classification. In particular, two undirected graphs are used to model the H channel and the S and I channels respectively in order to represent a color texture image. Moreover, the shortest paths is constructed by using four pairs of pixels according to different scales and directions of the texture image. Experimental results on colored Brodatz and USPTex databases reveal that our proposed method is effective, and the highest classification accuracy rate is 96.93$\%$ in the Brodatz database.
\end{abstract}

\section{Introduction}
Texture analysis is an active research topic in computer vision and pattern recognition. Its application is very extensive, including texture classification, segmentation, synthesis and retrieval. A texture is generally defined as a complex visual pattern composed of entities£¬or sub-patterns£¬with specific size£¬brightness and slope etc. As an important step in texture classification, texture feature extraction is to extract a discriminative feature from these complex visual pattern. Many scholars have made contributions to feature extraction that is usually divided into four major categories: statistical, signal processing, structural and model based methods. Statistical based methods such as Haralick et al. used the co-occurrence matrices to model the texture pattern \cite {clar:13}. Signal processing methods is also called transform-based methods. The widely used transforms include Gabor transform \cite {clar:14}, wavelet transforms \cite {clar:15},\cite {clar:20},\cite {clar:23},\cite {clar:24}, contourlet transforms \cite {clar:21} and shearlet transforms \cite {clar:22}. The typical structural approaches are the local binary pattern-based methods including complete local binary pattern (CLBP) \cite {clar:16} and scale selective local binary pattern (SSLBP) \cite {clar:17} and so on. The model based methods that depend on stochastic models to interpret image texture. In addition, some scholars put forward other methods, such as Martinez et al. used deterministic tourist walk to analyze and classify texture \cite {clar:5}, Backes et al. proposed fractal descriptors \cite {clar:6} , complex network theory proposed by Costa et al. \cite {clar:7} and simplified gravitational systems \cite {clar:8}, \cite {clar:9}.\\\indent
However, most of these analysis methods are on gray-scale image. In the real world, color texture is the main form of existence. The recognition of color texture is more consistent with human vision. Therefore, color texture classification is still a challenge in the field of texture classification. In the past literature, we know that many scholars have done a lot of research on color texture analysis. For example, Drimbarean et al. extended the three gray-scale related texture analysis methods to color image \cite {clar:3}, Harvey et al. compared different color texture classification methods on the theoretical and experimental results \cite {clar:4}. Recently, Li et al. used Gaussian copula models of Gabor wavelets to analysis color texture \cite {clar:18}, Napoletano et al. compared hand-crafted and learned descriptors for color texture classification \cite {clar:19}.\\\indent
In this paper, we extend the analysis methods proposed in \cite {clar:1} and \cite {clar:2} respectively in RGB and HSV color space to HSI color space. Particularly, two undirected graphs are used to model the H channel and the S and I channel respectively in order to model a color texture image. First-order statistic of the shortest path is constructed by using four pairs of vertices according to different directions is calculated as features of the texture (called Shortest Paths in Graphs method $-$ SPG method).\\\indent
The remainder of the paper is organized as follows. Section 2 shows how a color texture can be modeled as graph in HSI color space. Section 3 presents the shortest path as the texture descriptors. In Subsection 4, we describe experiments in which our approach is compared against some traditional methods and the results achieve by our proposed methods. Finally, Section 5 gives a brief conclusion.\\
\section{Model Texture to Graph }
\subsection{Graph and Shortest Path}
In graph theory, a graph $G = (V, E)$ is composed of a vertex set $V$ and an edge set $E$. Due to that an digital image is expressed by its pixels, it is reasonable that any pixel in a given image texture can be considered as a vertex and relationship between any pair of pixels can be represented by an edge. As a classical problem in graph theory, the shortest path problem aims at finding the minimum path form the initial vertex to target vertex in a given weighted directed graph or undirected graph.
\subsection{Modeling Texture to Graph}
\subsubsection{Modeling an Undirected Graph from Texture.}
Graphs can be divided into directed graphs and undirected graphs. The difference between them is whether they are directional. In undirected graphs, the edge of the link vertex $v_{i}$ and $v_{j}$ lacks orientation, $(v_{i}, v_{j}) = (v_{j}, v_{i})$. On the other hand, there is a clear starting vertex and end vertex, $(v_{i}, v_{j}) \ne (v_{j}, v_{i})$. We aim to build a discriminant feature that characterizes an input texture image based on its undirected graph representation, which the first step of this processing is constructing the undirected graph that represents the neighborhood relation of the given image. A graph $G=(V, E)$ is composed of the set of vertices $V$ and edges $E$. We first set up the vertices set $V$. In this model, each pixel $I (x, y), x =1, ¡­,M$ and $y = 1, ¡­, N$ is viewed as a vertex belong to the set. The location of this vertex in the graph is the same as in the original image. Secondly, we propose to represent the neighborhood relationship by using the edges set $E$, connecting to each pair of vertices where the Chebyshev distance between them is shorter than or equal to a threshold value $t$ ($t$=1 in our paper):
\begin{equation}
E = \left\{ {e = \left( {v,v^0 } \right) \in V*V|\max \left( {\left| {x - x_0 } \right|,\left| {y - y_0 } \right|} \right) \le 1} \right\}
\end{equation}
where $x$ and $y$ are the Cartesian coordinates of the pixel $I(x, y)$ associated to the vertex $v$. Finally, weight $w(e)$ assigned to each $e \in E$, which is defined as:
\begin{equation}
w(e) = \left| {I\left( {x,y} \right) - I\left( {x_0 ,y_0 } \right)} \right| + \frac{1}{2}\left[ {I\left( {x,y} \right) + I\left( {x_0 ,y_0 } \right)} \right]
\end{equation}
where $I(x, y) = g, g=0,\dots , L$ represents the intensity value of the corresponding pixel. Figure 1 shows an undirected graph on a gray-scale image.
\begin{figure}
\centerline{\includegraphics[width=0.95\textwidth]{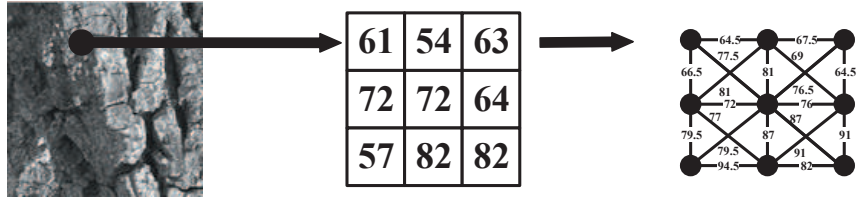}}
\caption{Undirected graph on gray-scale image}
\end{figure}
\subsubsection{Graph Based Representation of an HSI Color Texture.}
HSI color space is similar to RGB color space, each color is composed of three color components. But the difference is that the three color components of HSI are Hue (H), Saturation (S), and Intensity (I), where H represents the attribute of the pure color, S measures the extent of pure being diluted by the white light, and I is a subject description and a key parameter for human perception of color. The difference between the HSI color space and the HSV color space includes the difference in the representation of the first model, and second in the HSV, the component Value (V) represents the degree of bright color, and finally the calculation of the component V and the component I form the RGB color space is different.\\\indent
In the process of building a graph. First, we regard pixels in each color channel as a vertex, and connect each pixel in a specific channel. This process is similar to the color texture analysis method introduced in article [1], [2]. Let $I_{H}(x, y)$, $I_{S}(x, y)$ and $I_{I}(x, y)$ be each color component in an HSI color texture. According to the method described in the previous section, each color components is constructed into an undirected graph. In this way, three separate undirected graphs in HSI color space
represent a color texture image.\\\indent
Furthermore, we would explore the neighborhood relationship between the color components belonging to neighbor pixels. Because the HSI color space reflects the way people perceive the color of the visual system. Human vision is most intuitive to hue perception, we suggest that the H channel be regarded as an independent undirected graph to represent the attribute of the pure color, and the other two channels S and I can be connected to construct a new graph. By adding restrictions to Equation (1), we create the edge that connects the S channel and I channel. The condition is $\left \{e =(v, v_{0}) | v \in S \wedge v_{0} \in I \right \}$. As shown in Figure 2, texture images are represented by two undirected graphs.

\begin{figure*}[!ht]
\centering
\begin{minipage}[t]{0.45\textwidth}%²¢ÅÅ·ÅÁ½ÕÅͼƬ£¬Ã¿ÕÅÕ¼Ò³ÃæµÄ0.5£¬ÏÂͬ¡£
\centering
\includegraphics[width=0.90\textwidth]{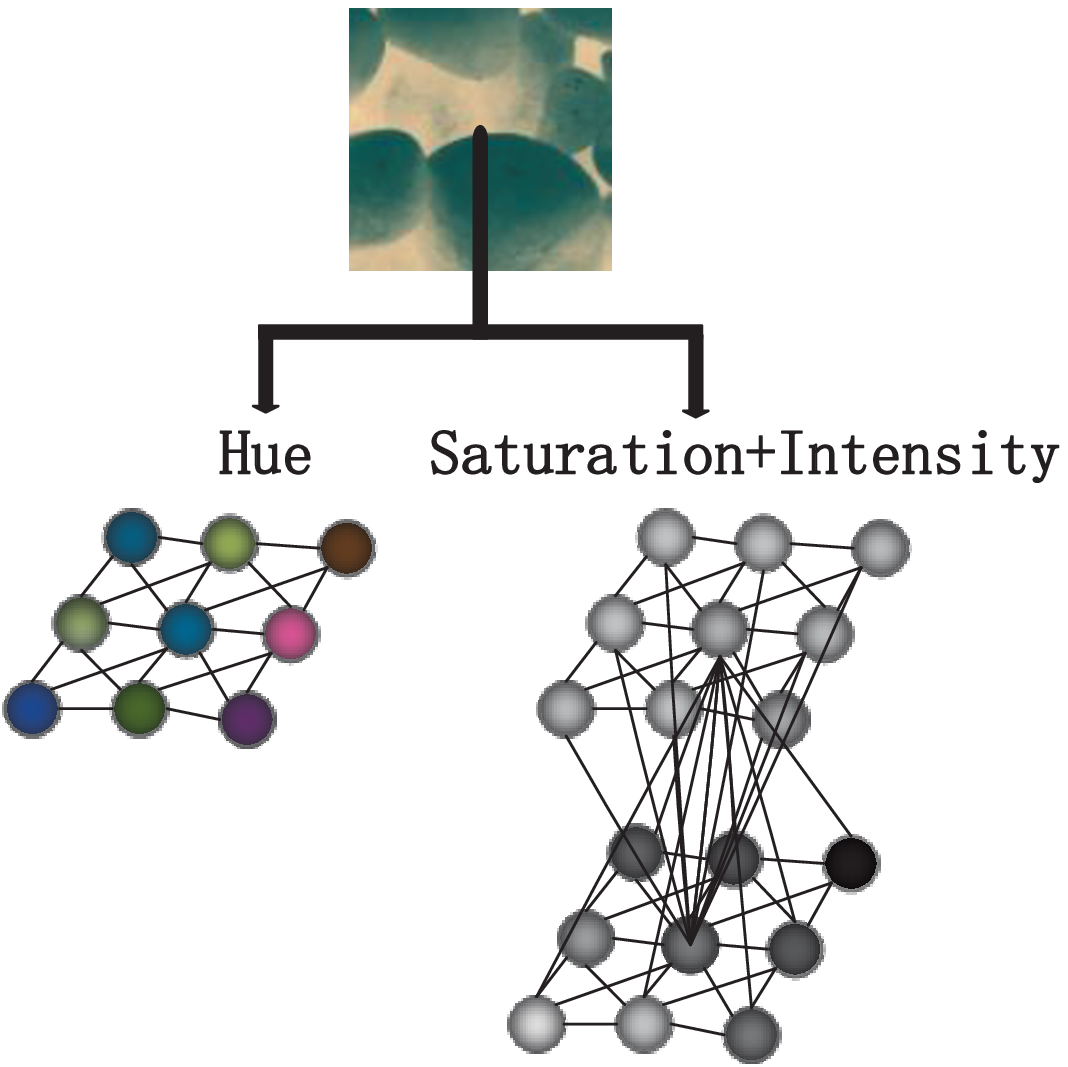}
\caption{Modeling a HSI color texture}%×¢ÊÍ1.jpg
\label{tet}
\end{minipage}
\begin{minipage}[t]{0.45\textwidth}
\centering
\includegraphics[width=0.90\textwidth]{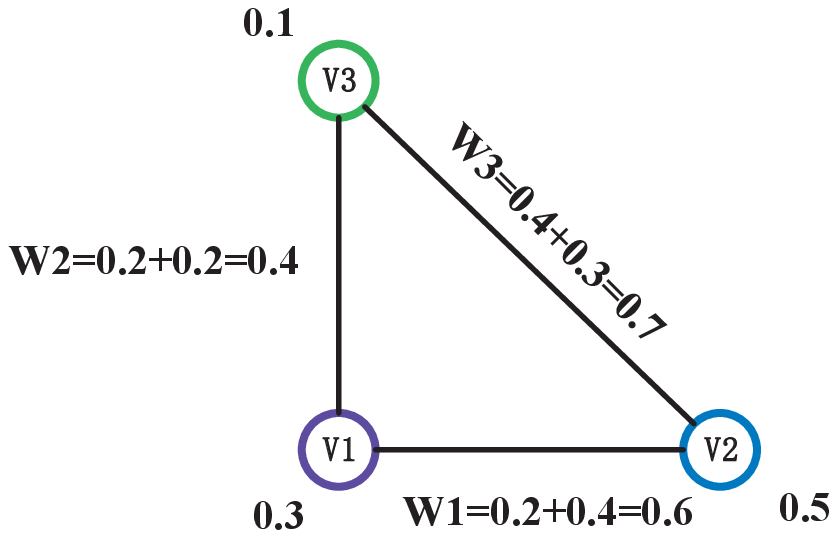}
\caption{Weight equation analysis}
\label{toc}
\end{minipage}
\end{figure*}
\subsection{Edge Weight Analysis}
In order to calculate the shortest path, Equation (2) is used to assign weights to each side of the undirected graph. The calculation of weights consists of two parts,each of which plays an important role in extracting feature vectors.\\\indent
The first part of Equation (2) is the difference between the intensities of two color components. We use the absolute value of the difference to avoid negative values of weights. As Figure 3 shows above, exploring the shortest path from vertex V3 that maintains the intensity of the 0.1 color component, the shortest path will choose the edge between vertex $V3$ and $V1$. Obviously, the absolute value of the difference between the vertex $V3$ and $V1$ is smaller than the value of between vertex $V3$ and $V2$. This means that those two vertices are similar.\\\indent
Moreover, the second part of Equation (2) shows the average intensity between two vertices. In Figure 3, the absolute value of the difference between the vertices $V3$ and $V1$ is equal to the absolute value of the difference between the vertices $V2$ and $V1$. We have to consider the average intensity between two vertices as a second part of the weight equation. This part emphasizes exploring the lower levels of intensity in the texture image. By considering the two parts, we can obtain a balance absolute value of the difference and average intensity minimization during the computes of the shortest path in the image.\\\indent
It is worth noting that in the undirected graph jointly represented by the S channel and I channel, the starting point and terminal point of the shortest path must be located in the same channel.
\section{Express Color Texture Features with the Shortest Path}
As mentioned above, we propose to construct two undirected graphs to represent the color texture images. One is an undirected graph composed of H-channel, and the other is an undirected graph composed of S-channel and I-channel. The shortest paths on two undirected graph are calculated as a feature vector. In order to hold as much as possible of information about the color texture, we recommend to calculate the shortest path between the four sets of vertices in different directions. The paths in these four directions are the diagonal direction (path $p_{45^\circ}$ and $p_{-45^\circ}$), horizontal (path $p_{0^\circ}$) direction and vertical path (path $p_{90^\circ}$), as shown in Figure 4.\\\indent

\begin{figure*}[!ht]
\centering
\begin{minipage}[t]{0.45\textwidth}%²¢ÅÅ·ÅÁ½ÕÅͼƬ£¬Ã¿ÕÅÕ¼Ò³ÃæµÄ0.5£¬ÏÂͬ¡£
\centering
\includegraphics[width=0.70\textwidth]{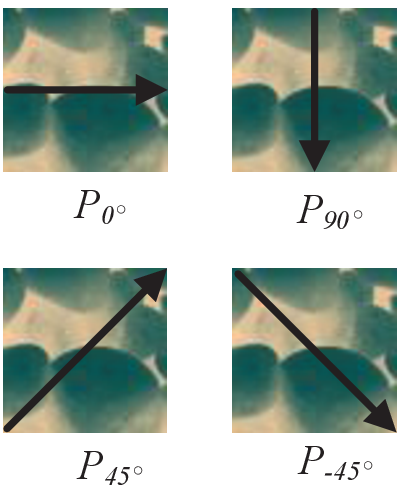}
\caption{Four directions of the shortest paths}%×¢ÊÍ1.jpg
\label{tet}
\end{minipage}
\begin{minipage}[t]{0.5\textwidth}
\centering
\includegraphics[width=0.8\textwidth]{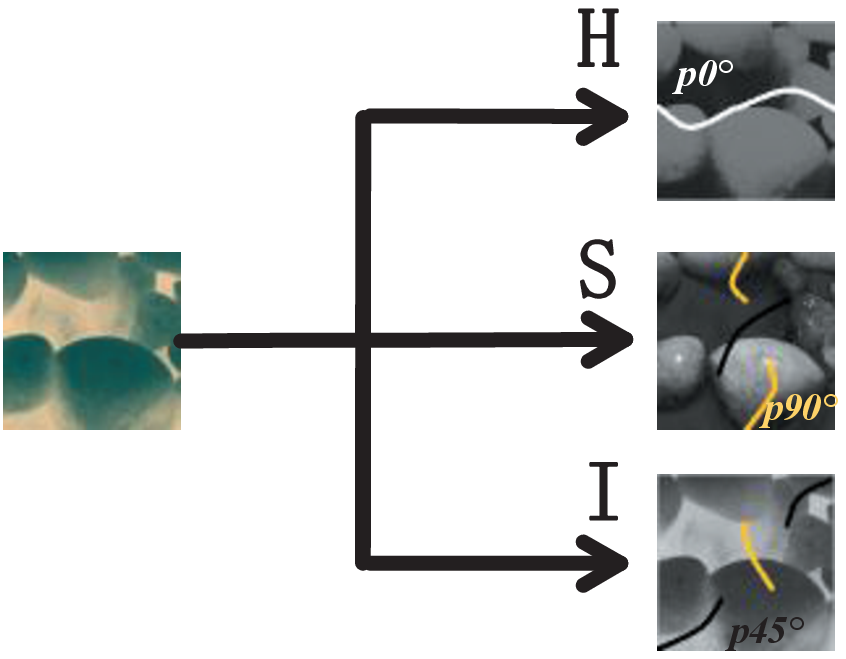}
\caption{Examples of shortest path}
\label{toc}
\end{minipage}
\end{figure*}
The starting and ending vertices of the shortest path are respectively on the three color components. We can obtain a total of twelve paths. On the other hand, as both saturation and intensity correspond to one single undirected graph, the search for the shortest path according to a specified orientation will indeed explore both color channels as shown in Figure 5.\\\indent
In order to describe the local texture image information as much as possible, we suggest covering the texture with a grid of size $r*r$, where $r$ is a divisor of the original texture size. For each grid block, we calculate the four shortest paths $(p_{0^\circ}, p_{45^\circ}, p_{-45^\circ}, p_{90^\circ})$ on the three color components (for example, we can cover an image 128*128 pixels with 32*32 grid, each grid block size is 4*4 pixels). We use the Dijkstra¡¯s algorithm to find the shortest path. Then, for each path direction $d^\circ$, $d^\circ = \left \{0^\circ, 45^\circ, -45^\circ, 90^\circ \right \}$, we compute the first-order statistics in each direction as average $\mu _{d^\circ}$ and standard deviation $\sigma _{d^\circ}$. By following this strategy, the feature vector can be defined as:
\[
\vec{\alpha} _r^c  = \left[ {\mu _{0^{\rm ^\circ } } ,\sigma _{0^{\rm ^\circ } } ,\mu _{45^{\rm ^\circ } } ,\sigma _{45^{\rm ^\circ } } ,\mu _{ - 45^{\rm ^\circ } } ,\sigma _{ - 45^{\rm ^\circ } } ,\mu _{90^{\rm ^\circ } } ,\sigma _{90^{\rm ^\circ } } } \right]
\]
where $r$ is the grid size of covering texture in a specific color channel $C = \left \{H, S, I\right \}$. Thus, one single vector that holds all the texture image texture is composed of these vectors $\vec{\alpha}_r^H$,$\vec{\alpha}_r^S$ and $\vec{\alpha}_r^I$, which are the feature vectors characterizing each of the color channels:
\[
\vec \theta _r  = \left[ {\vec \alpha _r^H ,\vec \alpha _r^S ,\vec \alpha _r^I } \right]
\]\\\indent
According to the different size of the grid scale covering the texture, a multi-scale analysis can be accomplished by concatenating previous feature vectors
corresponding to different grid size:

\[
\vec \theta _{r_1 ,r_2 ,...,r_n }  = \left[ {\vec \theta _{r_1 } ,\vec \theta _{r_2 } ,...,\vec \theta _{r_n } } \right]
\]

\section{Experiment and Results}
In this section, various experiments are carried out to demonstrate the efficiency of our proposed texture classification method. The parameter $r$ is used to define the grid size $r*r$ for covering texture image. In order to avoid dealing with overlapping grid, we choose the common divisors of the original texture image size (128*128 or 160*160) as the set of grid size $r$ values: $r$={4,8,16,32}. 1 values were omitted because it does not provide the desire information about the standard deviation, which makes the composition of the $\theta_{r}$ not possible. Moreover, when the grid size is 128*128 or 160*160, each grid block size is 1*1 pixel, it is too small to provide meaningful information about the texture.\\\indent
We use the one nearest neighbor (1NN) classifier to evaluate the accuracy of our proposed approach for discrimination. At the same time, we use leave-one-out cross-validation that is using each sample in the database for validation while the remaining samples are used as the training set. And we also use holdout strategy by considering 2/3 of the samples for training and 1/3 for testing, with 10 repetitions.\\\indent
In this experiment, we evaluate our proposed texture classification on the colored Brodatz database and USPTex. Colored Brodatz not only preserves the advantages of the original Brodatz database rich texture content, but also has a variety of color content. For the experiment using this database, we consider a total of 1792 samples of 160*160 pixels size grouped in to 112 texture classes (16 texture samples per class). USPTex database consists of common textures, we consider a total 2292 samples of 128*128 pixels size grouped into 191 texture classes (12 texture samples per class). \\\indent
%\begin{figure}
%\centerline{\includegraphics[width=0.95\textwidth]{Fig6.eps}}
%\caption{Examples of each texture class in USPTex database}
%\end{figure}\\\indent
The proposed method is evaluated by two types of analysis, namely single-scale and multi-scale. Firstly, we conduct single-scale analysis. As seen in the Table 1, our proposed method has achieved good results in the Brodatz, the highest accuracy can reach 96.93$\%$ and 97.38$\%$. As the size of the grid increase, so does the accuracy. This is due to the fact that the increase in grid size corresponds to a decrease in the size of the grid blocks to ensure better local texture analysis performance.\\\indent
The second time we conduct a multi-scale analysis, the experimental results were slightly lower than single-scale. Table 2 shows the multi-scale results.
\begin{table}
\caption{The results of a single-scale on two texture database}
\begin{center}
\begin{tabular}{rrrrrcccccccccccccccccccccccc}
\hline
           &            &            &            &            &                                                                                                                                                                                                                                                                                    \multicolumn{ 24}{c}{Accuracy(\%)} \\
\hline
           &            &            &            &            &                                                                                               \multicolumn{ 10}{c}{1NN+holdout} &            &            &            &            &                                                                                         \multicolumn{ 10}{c}{1NN+leave-one-out} \\
\hline
 Grid size &            &            &            &            &      32*32 &            &            &      16*16 &            &            &        8*8 &            &            &        4*4 &            &            &            &            &      32*32 &            &            &      16*16 &            &            &        8*8 &            &            &        4*4 \\

   Brodatz &            &            &            &            &      96.93 &            &            &      90.77 &            &            &       86.2 &            &            &      73.38 &            &            &            &            &      97.38 &            &            &       91.8 &            &            &      88.06 &            &            &      75.17 \\

    USPTex &            &            &            &            &       57.8 &            &            &      31.13 &            &            &      28.18 &            &            &      24.06 &            &            &            &            &      59.81 &            &            &      33.03 &            &            &      29.93 &            &            &      24.04 \\
\hline
\end{tabular}
\end{center}
\end{table}
\begin{table}
\caption{The results of a multi-scale on two texture database}
\begin{center}
\begin{tabular}{ccccccccccccccccccc}
\hline
           &            &            &            &            &            &                                                                                                                                    \multicolumn{ 13}{c}{Accuracy (\%)} \\
\hline
           &            &            &            &            &            &                                                                                                                                      \multicolumn{ 13}{c}{1NN+holdout} \\
\hline
Multi-scale &            &            &            &            &            &    \{32,16\} &            &            &            &            &            &  \{32,16,8\} &            &            &            &            &            & \{32,16,8,4\} \\

   Brodatz &            &            &            &            &            &      96.46 &            &            &            &            &            &      95.41 &            &            &            &            &            &      54.75 \\

    USPTex &            &            &            &            &            &      54.75 &            &            &            &            &            &       50.9 &            &            &            &            &            &      45.55 \\
\hline
           &            &            &            &            &            &                                                                                                                                \multicolumn{ 13}{c}{1NN+leave-one-out} \\
\hline
   Brodatz &            &            &            &            &            &      96.99 &            &            &            &            &            &      96.14 &            &            &            &            &            &      94.81 \\

    USPTex &            &            &            &            &            &      56.46 &            &            &            &            &            &      52.53 &            &            &            &            &            &      47.21 \\
\hline
\end{tabular}
\end{center}
\end{table}\\\indent
Finally, we compare our proposed method with other representative texture classification methods including single-band SPG in RGB \cite{clar:1}, HRF \cite{clar:10}, MultiLayer CCR \cite{clar:11} and MSD \cite{clar:12} by the considering the feature vector $\vec \theta _{4}$ with 1NN + holdout on USPTex database. In particular, we compared with the single-band methods in the RGB color space introduced in the literature \cite{clar:1}, all of which are the same size grid (32*32, each grid block size is 4*4 pixels). Additionally, we also tested our approach against a simple method of average values of each R, G and B. The results shows in Table 3.
\begin{table}
\caption{Compared with other classic method on USPTex}
\begin{center}
\begin{tabular}{ccccccccccc}
\hline
           &            &            &            &            &            &            &            &            &            & Accuracy (\%) \\
\hline
    Method &            &            &            &            &            &            &            &            &            & 1NN+holdout \\
\hline
Propose method &            &            &            &            &            &            &            &            &            &       57.80 \\

Single-band SPG in RGB \cite {clar:1} &            &            &            &            &            &            &            &            &            &      54.39 \\

  HRF \cite {clar:10} &            &            &            &            &            &            &            &            &            &      49.86 \\

MultiLayer CCR \cite {clar:11} &            &            &            &            &            &            &            &            &            &      82.08 \\

  MSD \cite {clar:12} &            &            &            &            &            &            &            &            &            &      51.29 \\

Average RGB &            &            &            &            &            &            &            &            &            &      36.19 \\
\hline
\end{tabular}
\end{center}
\end{table}
\section{Conclusion}
In this paper, we propose to use shortest paths in the HSI space to build a texture representation for classification. Two undirected graphs is used to model texture
images, which are undirected graphs constructed by H channel and graphs constructed jointly by S channel and I channel. A texture representation vector can be obtained by computing the shortest path in four different directions in a specific channel. Experimental results show that our proposed color texture classification method is effective.\\\indent
\section*{Acknowledgements}
This work was supported in part by Program for Science $\&$ Technology Innovation Talents in Universities of Henan Province under Grant 19HASTIT026, and in part by the Training Program for the Young-Backbone Teachers in Universities of Henan Province under Grant 2017GGJS065.

\end{document}